\documentclass{article}

\usepackage[final]{nips_2016}

\usepackage[utf8]{inputenc} % allow utf-8 input
\usepackage[T1]{fontenc}    % use 8-bit T1 fonts
\usepackage{hyperref}       % hyperlinks
\usepackage{url}            % simple URL typesetting
\usepackage{booktabs}       % professional-quality tables
\usepackage{amsfonts}       % blackboard math symbols
\usepackage{nicefrac}       % compact symbols for 1/2, etc.
\usepackage{microtype}      % microtypography

\usepackage{todonotes}
\usepackage{color,soul}

\usepackage{algpseudocode}
\usepackage{algorithm}
\usepackage{amsmath}
\usepackage{amssymb}
\usepackage{bm}
\usepackage{subfig}
\usepackage{tikz}
\usepackage{mathtools}

\tikzstyle{block} = [rectangle, draw, 
    text width=8em, text centered, rounded corners, minimum height=4em]
\tikzstyle{line} = [draw, -latex]

\newtheorem{definition}{Definition}[section]
\DeclareMathOperator*{\E}{\mathbb{E}}
\DeclarePairedDelimiter\Bracket{\big[}{\big]}

\title{DQN with Model-Based Exploration: Efficient Learning on Environments with Sparse Rewards}

\author{
	Stephen Gou \\
	Department of Computer Science\\
	University of Toronto\\
	\text{gouzhen1@cs.toronto.edu} \\
	\And
	Yuyang Liu \\
	Department of Computer Science\\
	University of Toronto \\
	\texttt{\href{mailto:yuyang@cs.toronto.edu}{yuyang@cs.toronto.edu}} \\
}

\begin{document}

\maketitle

\begin{abstract}
  We propose Deep Q-Networks (DQN) with model-based exploration, an algorithm combining both model-free and model-based approaches that explores better and learns environments with sparse rewards more efficiently. DQN is a general-purpose, model-free algorithm and has been proven to perform well in a variety of tasks including Atari 2600 games since it's first proposed by Minh et el\cite{Mnih2015}. However, like many other reinforcement learning (RL) algorithms, DQN suffers from poor sample efficiency when rewards are sparse in an environment. As a result, most of the transitions stored in the replay memory have no informative reward signal, and provide limited value to the convergence and training of the Q-Network. However, one insight is that these transitions can be used to learn the dynamics of the environment as a supervised learning problem. The transitions also provide information of the distribution of visited states. Our algorithm utilizes these two observations to perform a one-step planning during exploration to pick an action that leads to states least likely to be seen, thus improving the performance of exploration. We demonstrate our agent's performance in two classic environments with sparse rewards in OpenAI gym\cite{openai-gym}: Mountain Car and Lunar Lander.
\end{abstract}

\section{Introduction}
Reinforcement learning agent learns by interacting with the environment and uses observed reward for each action as feedback signal to improve policy. In some environments, there are constant reward signals. For example, the score of the game when training an agent to play Pong (Atari game) or the distance travelled when training a robot to run. In such environments, the agent continuously receives constructive reward feedback, providing strong signals and gradients to train the agent's underlying model.

However, in other environments, desired outcomes are rare, and the agent only receives a reward when the desired outcome happens. For instance, in the Atari game Montezuma’s Revenge, the agent only receives a reward for picking up a key that requires performing a series of tasks successfully. The agent can only start improving the model when it accidentally stumbles into one successful sequence of actions by random actions. Given the extremely low probability, it usually requires extremely large number of training episodes, especially in the beginning, which could be very costly in real world environments.

One way to combat this problem is to design algorithms that can explore environments faster and more thoroughly. In DQN, the agent typically uses a $\epsilon$-greedy policy to decide exploitation or exploration, and chooses a random action during exploration, which is extremely inefficient in environments with sparse rewards. Therefore, we propose an improved version of DQN that performs a one-step planning during exploration, increasing the chance of discovering unseen states.

\section{Background}
\label{background}
Unlike supervised and unsupervised learning, which involve learning from data given upfront, reinforcement learning tries to retro-feed its model by observing rewards through interactions with the environment in order to improve. Delayed rewards and interactions with the underlying environments are the two major characteristics of reinforcement learning \cite{Sutton:2018:IRL}.

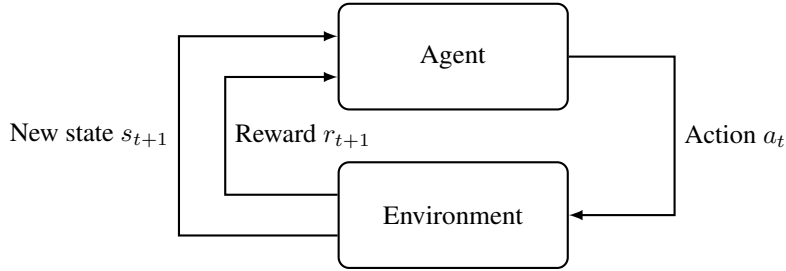
\begin{figure}[ht]
  \centering
  \begin{tikzpicture}[node distance = 6em, auto, thick]
    \node [block] (Agent) {Agent};
    \node [block, below of=Agent] (Environment) {Environment};
    
     \path [line] (Agent.0) --++ (4em,0em) |- node [near start]{Action $a_t$} (Environment.0);
     \path [line] (Environment.190) --++ (-6em,0em) |- node [near start] {New state  $s_{t+1}$} (Agent.170);
     \path [line] (Environment.170) --++ (-4.25em,0em) |- node [near start, right] {Reward $r_{t+1}$} (Agent.190);
\end{tikzpicture}
  \caption{Agent interacts with the environment}
  \label{fig:MDP}
\end{figure}

Reinforcement learning consists of a sequence of interactions with the environment through actions and observes the reward and next state, illustrated by Figure \ref{fig:MDP}. This process can be formally defined as a Markov decision processe (MDP). 
\begin{definition}[Markov decision processes]
\quad 
Defined by: ($\mathcal{S}$, $\mathcal{A}$, $\mathcal{R}$, $\mathbb{P}$, $\gamma$), and a policy $\mathbf{\pi}$ \\
$\mathcal{S}$: set of possible states \\
$\mathcal{A}$: set of possible actions \\
$\mathcal{R}$: distribution of reward given state and action pair \\
$\mathbb{P}$: transition probability \\
$\gamma$: discount factor \\
$\mathbf{\pi}$: a function from $\mathcal{S}$ to $\mathcal{A}$ that tells which action to take in each state
\end{definition}

However, in many environments, the underlying dynamics, e.g. the transition probability is not known. Algorithms that can learn without knowing the dynamics are called model-free, and there are two main approaches: Q-learning and policy gradient.

\subsection{Q-Learning}
In Q-learning, the agent learns a Q-Value function that gives the expected total return given a state and action pair. At each time step, the agent acts with a greedy policy $\pi$, picking an action that maximizes the Q function.
\begin{definition}[Q-Value function]
\[
    Q^\pi(s,a) = \E \Bracket{\sum_{t\ge0} \gamma^t r_t | s_0=s, a_0=a, \pi}
\]
\end{definition}

The Q-Value function($Q^*(s,a)$) given a state and action pair is optimal when the agent uses the greedy policy. The Q-Value function involves the expectation of return over all future time steps, which is hard to learn. One can apply the Bellman operator to convert the equation into a recursive one:\\
\[
  Q^*(s,a) = \E_{s' \thicksim \mathcal{\varepsilon}} \Bracket{r + \gamma \max_{a'} Q^*(s',a')|s,a} 
\]
We can then apply the Value iteration algorithm to get an iterative update formula to learn the Q-values:
\[
  Q_{i+1}(s,a) = r(s,a) + \gamma \sum_{s' \in S}\mathcal{P}(s'|s,a) \max_{a'} Q_i(s',a')
\]
This method works fine if the state space and action space are relatively small where one can use a table to keep track of all the state-action pairs. However, when state action space becomes large, it is in-feasible to calculate this optimal Q-Value function exactly. Thus in Q-Learning, we use a function approximator instead.

\textbf{Deep Q-Network (DQN)}\\
DQN \cite{Mnih2015} uses a neural network, which can be a deep convolutional network if dealing with high dimensional state space like pixels, to approximate the Q-value function.
During each training step, the transition $(s_t, a_t, r_t, s_{t+1})$ is saved in an experience replay memory, and draws samples from it to train the network, increasing sample efficiency. It also deploys another target Q-network to provide Q-value estimates. The target network only gets updated every number of steps, increasing the stability of training.

\subsection{Policy Gradients}
Policy gradient methods directly learn the optimal actions without learning the values of states. The simplest policy gradient method is REINFORCE, also known as Monte Carlo Policy Gradient \cite{Sutton:1999:PGM:3009657.3009806}, is described below.

Given a set of all policies $\Pi = \{ \pi_\theta, \theta \in \mathbb{R}^m \}$, the expected return of a policy is defined as
\[
    J(\theta) = \E \Bracket{ \sum_{t \ge 0} \gamma^t r_t | \pi_\theta } = \E_{\tau \thicksim p(\tau;\theta)} \Bracket{r(\tau)} = \int_\tau r(\tau)p(\tau;\theta)d\tau
\]
where $\tau$ is the sequence of the trajectory$(s_0,a_o,r_0,s_1,\dots)$

The gradient of $J(\theta)$ is:
\begin{align*}
    \nabla_\theta J(\theta) 
    & = \int_\tau r(\tau) \nabla_\theta p(\tau;\theta)d\tau = \int_\tau (r(\tau) \nabla_\theta log p(\tau;\theta)) p(\tau;\theta) d\tau \\
    & = \E_{\tau \thicksim p(\tau;\theta)}\Bracket{ r(\tau) \nabla_\theta log p(\tau;\theta) }
\end{align*}
where we could use Monte Carlo estimate to find the gradient of $\nabla_\theta \, log \, p(\tau;\theta)$ :
\begin{align*}
    \nabla_\theta \, log \, p(\tau;\theta) 
    & = \nabla_\theta \, log \, \prod_{t \ge 0} p(s_{t+1} | s_t, a_t) \pi_\theta (a_t|s_t) \\
    & =  \sum_{t \ge 0} \nabla_\theta \, log \, \pi_\theta (a_t|s_t)\\
    \Rightarrow  \nabla_\theta J(\theta)  & =\E_{\tau \thicksim p(\tau;\theta)}\Bracket{ \sum_{t \ge 0} r(\tau) \nabla_\theta \, log \, \pi_\theta (a_t|s_t) }
\end{align*}

One can optimize a policy by performing gradient ascent of $J(\theta)$ with respect to $\theta$. The idea of relying on the reward of a particular trajectory can cause large variance during training, and one way to improve is combining Q-Learning and Policy Gradients, which is called Actor-Critic.

\section{Related Work}
\label{headings}

Improving exploration and learning efficiency of environments with sparse rewards is an active area of research. Our approach falls under the category of using heuristics as guidance to make an informed exploration step instead of picking a random action. Similar ideas have been presented before. In Oh et el.'s paper \cite{Oh2015} on predicting Atari games frames, their deep neural network architecture is able to generate next 100-step frames conditioned on actions with high accuracy. They use this information to guide exploration, choosing actions that will lead to rarer states. The rarity of a state comparing to recently visited states is computed by a Gaussian kernel. Similarly, Dilokthanakul et el.\cite{Dilokthanakul2018} proposed an improved exploration in DQN by informed exploratory actions that encourage visiting states whose values have high uncertainties. 

The use of intrinsic reward to provide feedback signal is another popular approach. For example, Pathak et el.\cite{Pathak2017} introduced curiosity-driven exploration, where it uses the error of the state prediction by a forward dynamics model against the true next state as an intrinsic reward. And the agent is trained to maximize the sum of the intrinsic reward plus environmental reward. 

Methods that improve sample-efficiency of RL algorithms are also helpful in environments with sparse rewards. For example, in the paper \textit{Prioritized Experienced Replay} \cite{Sachaul2015}, Schaul et el. improved DQN by sampling experience replays with priority instead of uniform sampling. The key observation was that transitions that are more surprising, less redundant and rarer provide more information for the agent to learn. They showed that increasing the sampling frequencies of these transitions result in faster learning. Azizzadenesheli et el.\cite{Azizzadenesheli2018} proposed a novel RL algorithm that combines both model-free and model-based methods to achieve better efficiency. They use a Generative Adversarial Network (GAN) to model the environment's dynamics as well as a predictor for reward. The algorithm utilizes these models to do planning by a Monte Carlo Tree Search (MCTS).

\section{DQN with Model-Based Exploration}

\label{others}

The full algorithm is presented in Algorithm 1. The agent chooses between exploration and exploitation based on an $\epsilon$-greedy policy. Like the original DQN algorithm, our agent trains two Q-networks, including a target Q-network to increase stability. Likewise, we utilize a replay memory and clip the error terms when training Q-network. On top of the DQN algorithm, we also train a dynamics network that predicts the next state given a state and action pair. Combining this dynamics network and an explicit modeling of the distribution of recently visited states, our agent is able to pick an action that increases the chance to visit unseen states during exploration. 
\subsection{Dynamics Network}
In environments with sparse rewards, most if not all of the transitions in replay memory have non-informative rewards, providing little signal for the agent to learn Q values. However, we utilize these transitions to train a neural network $D(s,a)_{\theta^D}$ that is able to predict  $s_{t+1}$ given current state $s_t$ and an action $a$. This network is crucial in making the guided exploration step. We use a fully connected feed-forward neural network (see Table \ref{table-dqn}). The dynamics network can be trained using the same transitions sampled from experience replay that is used to train the Q-network. Therefore, implementing a prioritized replay memory will benefit the training of the dynamics network as well.

\subsection{Guided Exploration}
\begin{figure}[htbp]
\centering
\subfloat[Random]{\scalebox{0.28}{\includegraphics{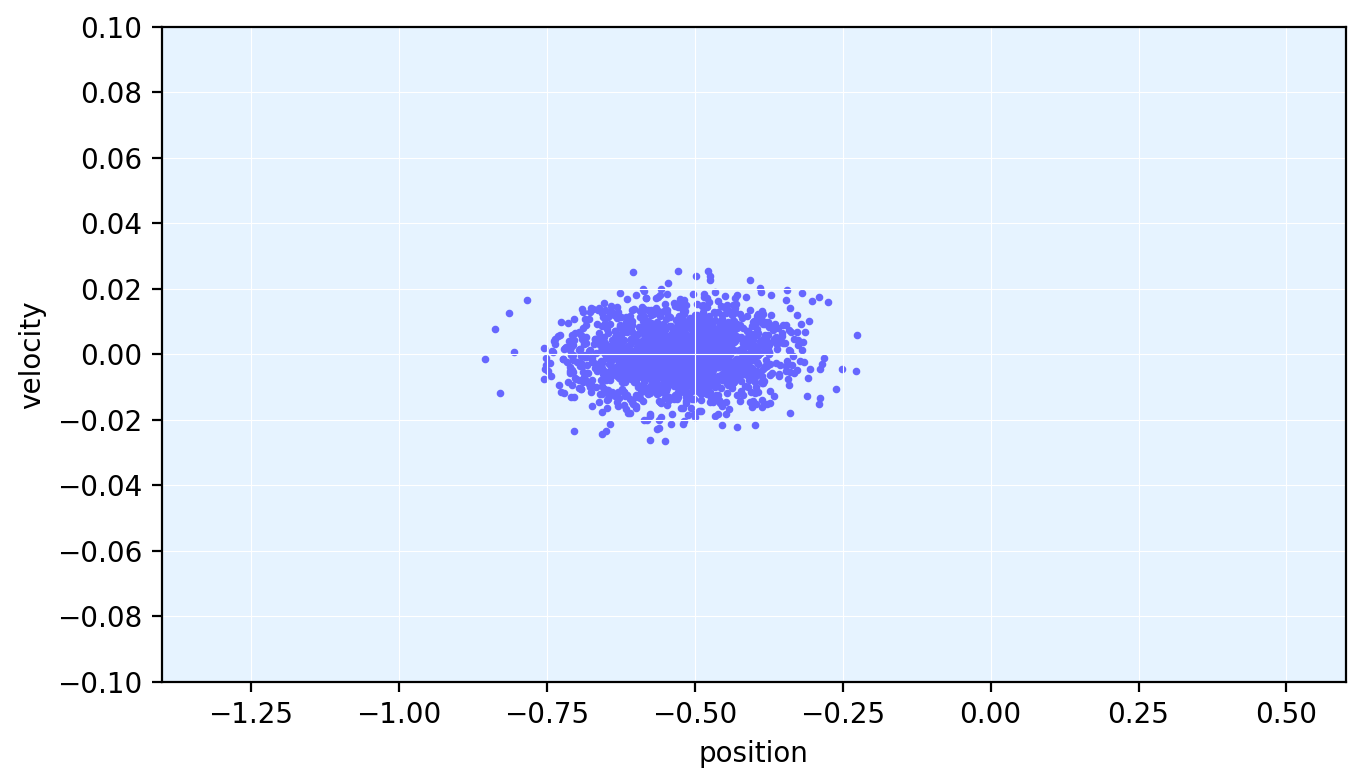}}\label{fig:f1}}
\hfill
\subfloat[Similarity by Gaussian kernel]{\scalebox{0.28}{\includegraphics{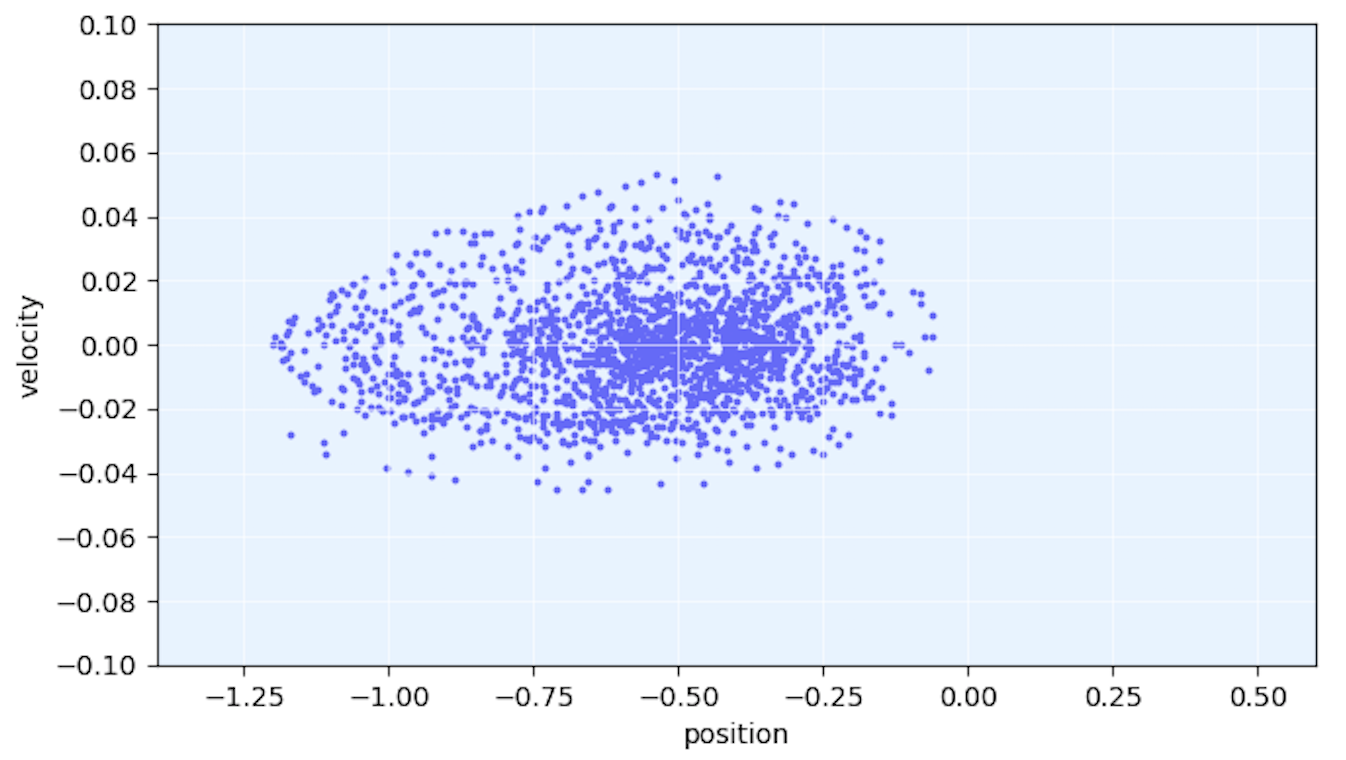}}\label{fig:f3}}
\hfill
\subfloat[Our Method]{\scalebox{0.28}{\includegraphics{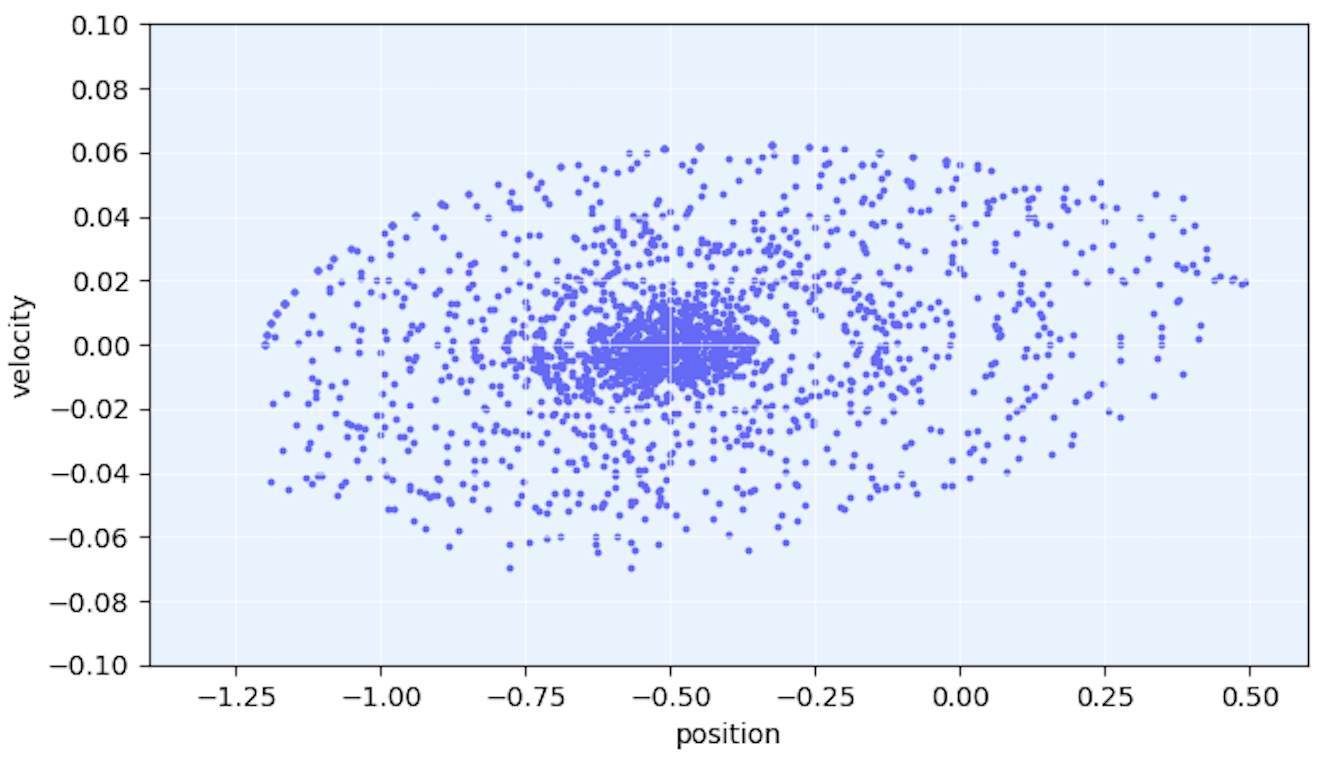}}\label{fig:f2}}
\caption[Comparing explored states in Mountain Car for different exploration methods]
{Scatter plots of explored states in Mountain Car after 50 episodes of only exploration. Our method was able to explore a wider range of states. The best result out of 3 independent runs is plotted for each method. \textbf{(a)} Explore with random action. \textbf{(b)} Pick an action that leads to a state which has the least similarity (measured by a Gaussian kernel) with recently visited states. \textbf{(c)} Our proposed algorithm for exploration.}
\label{mountain_state}
\end{figure}

The most common way to explore for an $\epsilon$-greedy policy is a uniform sampling in action space. However, as shown in Figure \ref{mountain_state} (a), using random actions to explore will result in: 1) most of the states concentrate around the initial state, 2) large area of the state space is never visited.\\

The goal of guided exploration is to utilize the learned dynamics of the environment to choose an action by a one-step planning during exploration such that there is a better chance of reaching rare or unseen states. At a given state, we can predict the next state for choosing each action in the action space, and we pick the action that leads to a state that is least similar comparing to the states we have seen. 

Unlike Oh et el. \cite{Oh2015} who uses a Gaussian kernel as similarity measure, we propose to evaluate the rarity of a state comparing to recently visited states $\mathcal{S}_F$ by a probabilistic approach. For simplicity and generality, we model the distribution of past states as a multivariate Gaussian with the empirical mean and empirical covariance of $\mathcal{S}_F$ as the parameters:\\
\[
\bm{s} \thicksim \mathcal{N}( \bm{s} | \bm{\mu_{\mathcal{S}_F}}, \bm{\Sigma_{\mathcal{S}_F}})
\]\\
We pick exploratory action that leads to a next state that has the lowest probability according to this distribution. Explicit modeling of past states as a multivariate distribution has two advantages: 1) it takes into account the correlation between dimensions of the state. For example, in the Mountain Car environment, a higher velocity is more common given that the car is at a higher position, 2) it considers the variance for each component, eliminating the need of normalization. As a result, our method provides better exploration comparing to measuring similarity between states simply by distance metrics.
\begin{algorithm}[htbp]
\caption{DQN with Model-Based Exploration}
\begin{algorithmic}[1]
\State Initialize replay memory M to capacity N
\State Initialize Q-network \textit{Q} with random weights $\theta$
\State Initialize target Q-network \textit{\^{Q}} with weights $\theta^{-} = \theta$
\State Initialize dynamics predictor \textit{D} with random weights $\theta^{D}$
\For{episode = 1, E}
\For{t=1,T}
\State Explore = True with probability $\epsilon$
\If {Explore}
\State Retrieve the last F states visited from transitions in M and store in $\mathcal{S}_F$
\State Compute mean $\bm{\mu_{\mathcal{S}_F}}$ and covariance $\bm{\Sigma_{\mathcal{S}_F}}$ of $\mathcal{S}_F$
\State Pick $a_t = \text{argmin}_a \text{ } \mathcal{N}(D(s_t,a; \theta^D) |\; \bm{\mu_{\mathcal{S}_F}},\bm{\Sigma_{\mathcal{S}_F}})$
\Else 
\State Pick $a_t = \text{argmax}_a Q(s_t,a; \theta) $
\State Execute $a_t \text{ and observe reward } r_t \text{ and new state } s_{t + 1}$
\State Store transition $(s_t,a_t,r_t,s_{t+1})$ in M
\State Sample a batch of transitions uniformly $(s_j,a_j,r_j,s_{j+1})$ from M
\State Set $y_j = r_j + \gamma \text{max}_{a^{'}}  \hat{Q} (s_j , a^{'}; \theta^- )$ if episode not done, else $y_j = r_j$
\State Perform gradient descent on $Q$ with respect to $\theta$ on $(y_j - Q(s_j,a_j;\theta))^2$
\State Perform gradient descent on $D$ with respect to $\theta^D$ on $(s_{j+1} - D(s_j,a_j;\theta^D))^2$
\State Every C Steps set $\hat{Q}$ = $Q$
\EndIf
\EndFor
\EndFor
\end{algorithmic}
\end{algorithm}

\section{Experiments}
We test the proposed algorithm on two classic simulated environments with sparse rewards: Mountain Car and Lunar Lander. We use \textit{OpenAI Gym}'s \cite{openai-gym} implementations (discrete actions version) of the two environments. \\

\textbf{1) Evaluate improvement on exploration}\\
We run our algorithm with only exploration and we visualize the states visited. We compare our result to two other exploration techniques: 1) random action 2) informed action by Gaussian Kernel similarity measure. Figure \ref{mountain_state} and Figure \ref{lunar_state} show the results for each environment respectively.\\

\begin{figure}[htbp]
\centering
\scalebox{0.3}{\includegraphics{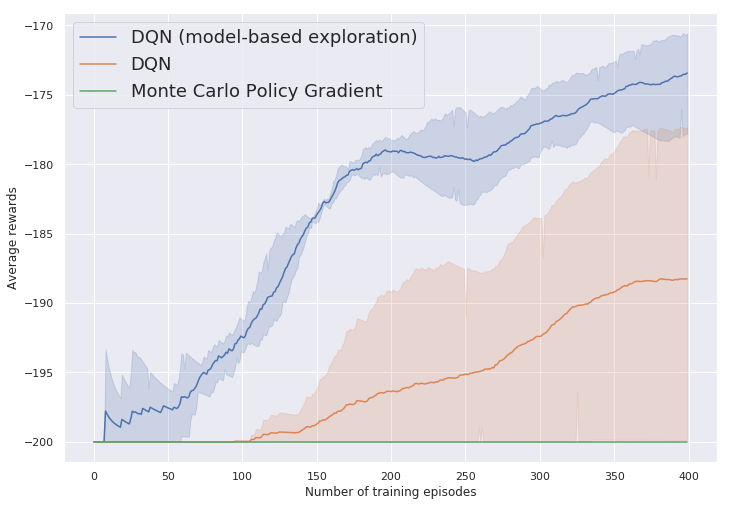}}
\caption[Running average reward in Mountain Car Envrionment]
{The running average rewards of 400 training episodes for Mountain Car, comparing our method against baselines. Our algorithm was able to improve rewards much sooner during training, comparing to the original DQN, which took at least 100 episodes to start seeing progress. Original DQN also failed to learn anything during one run, and Monte Carlo Policy Gradient always failed to learn anything in all runs. Each plot involves 3 independent runs of each algorithm. Solid lines represent the mean, and shaded areas represent range. }
\label{mountain_rewards}
\end{figure}

\begin{figure}
\centering
\scalebox{0.4}{\includegraphics{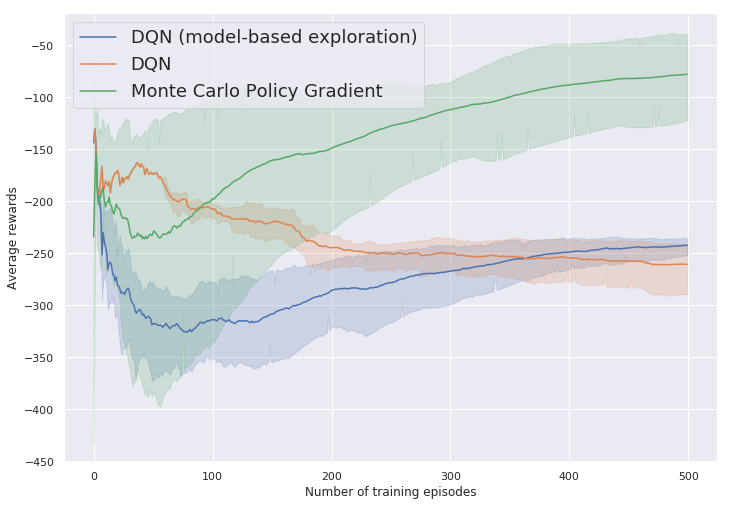}}
\caption[Running average reward in Lunar Lander Envrionment]
{The running average rewards of 500 training episodes for Lunar Lander, comparing our method against baselines. The policy gradient method achieved best result, while our method didn't show sign of improvement over original DQN. Each plot involves 3 independent runs of each algorithm. Solid lines represent the mean, and shaded areas represent range. }
\label{lunar_rewards}
\end{figure}

\textbf{2) Evaluate improvement on learning speed}\\
We evaluate the learning speed of our agent against two baselines: 1) original DQN, 2) Monte Carlo Policy Gradient. The running average of rewards for each environment is plotted in Figure \ref{mountain_rewards} and Figure \ref{lunar_rewards} respectively.

Our experiments showed that our proposed algorithm achieved significantly better exploration and learning speed in Mountain Car, but did not show any noticeable improvement in Lunar Lander.

\section{Limitations and Future Work}
Our proposed algorithm depends on two strong assumptions: 1) the dynamics of the environment can be learned with high accuracy, 2) the distribution of recently visited states follows a multivariate Gaussian distribution. Violation of either assumption can result in poor performance, which limits the application of our algorithm to certain environments.

\begin{figure}[htp]
\centering
\subfloat[Random]{\scalebox{0.28}{\includegraphics{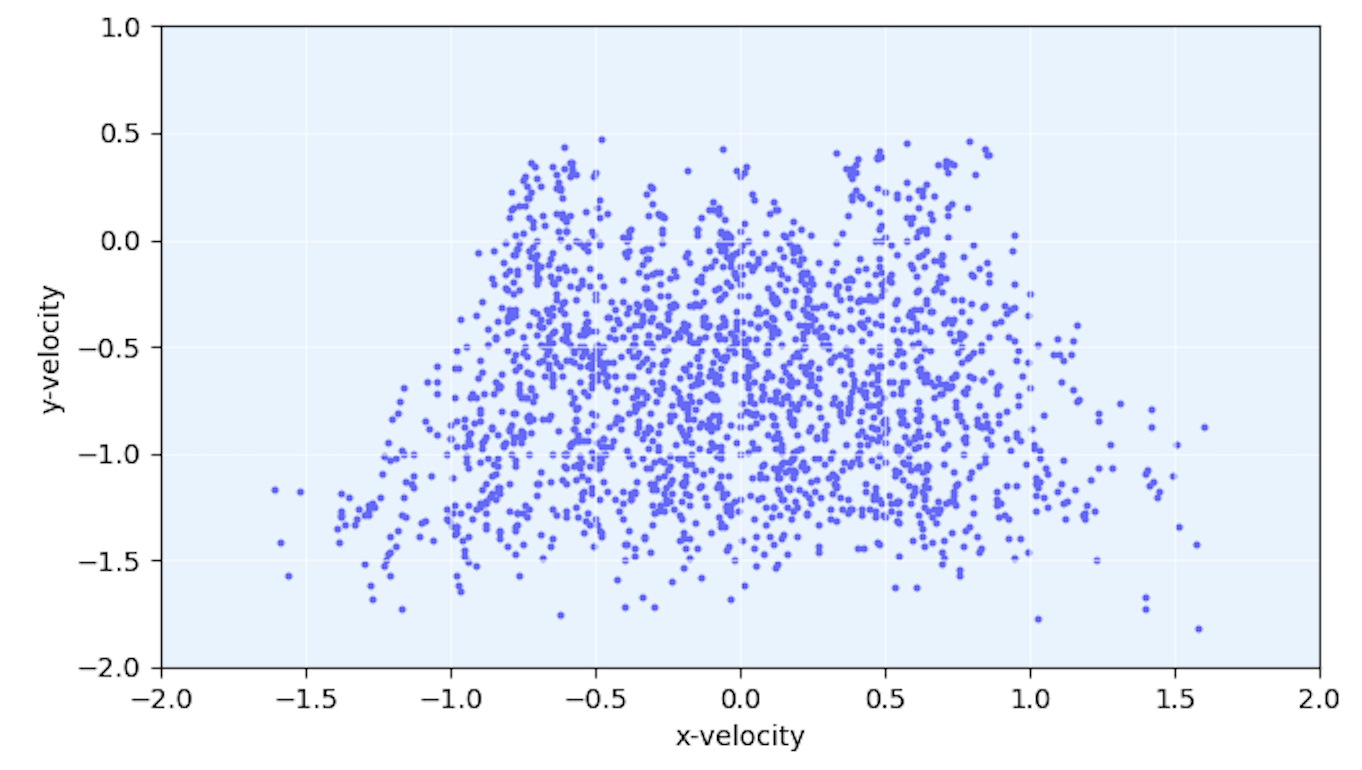}}\label{fig:f1}}
\hfill
\subfloat[Similarity by Gaussian kernel]{\scalebox{0.28}{\includegraphics{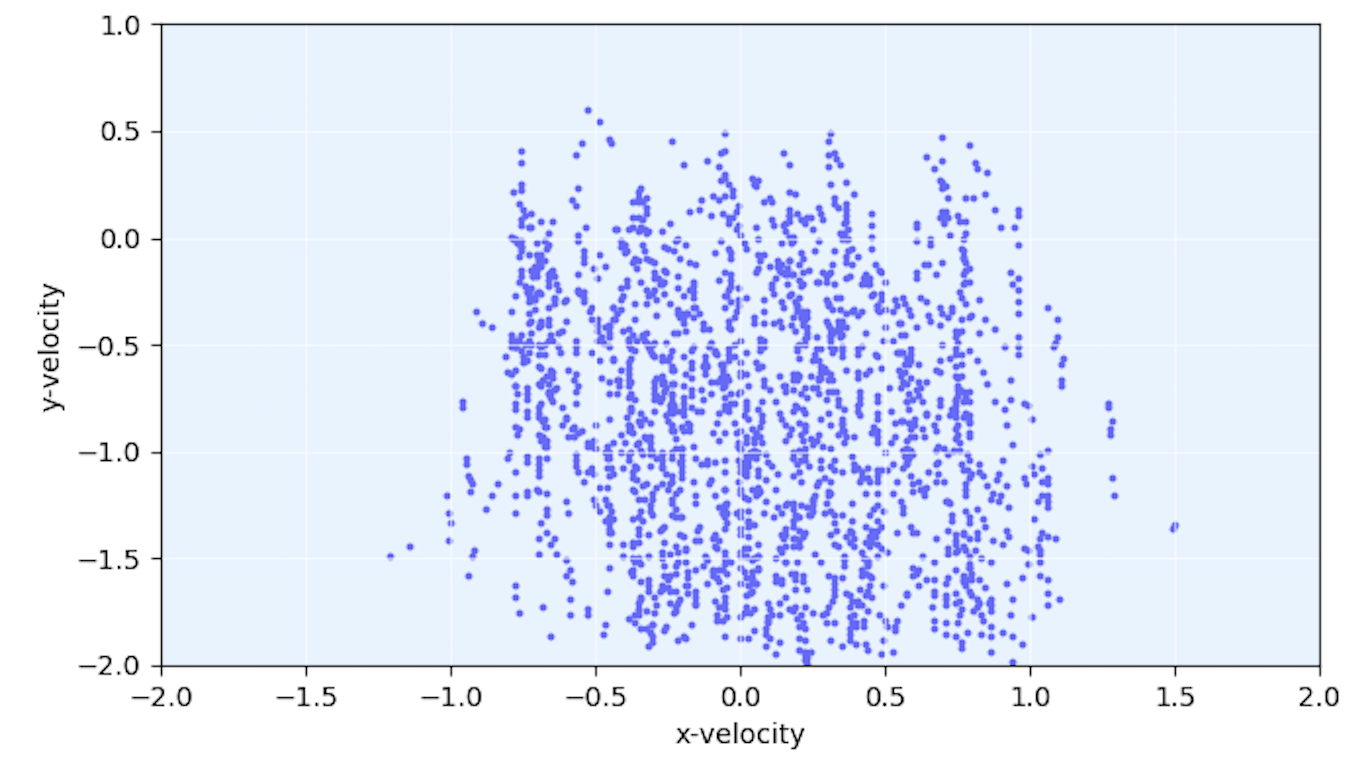}}\label{fig:f3}}
\hfill
\subfloat[Our Method]{\scalebox{0.28}{\includegraphics{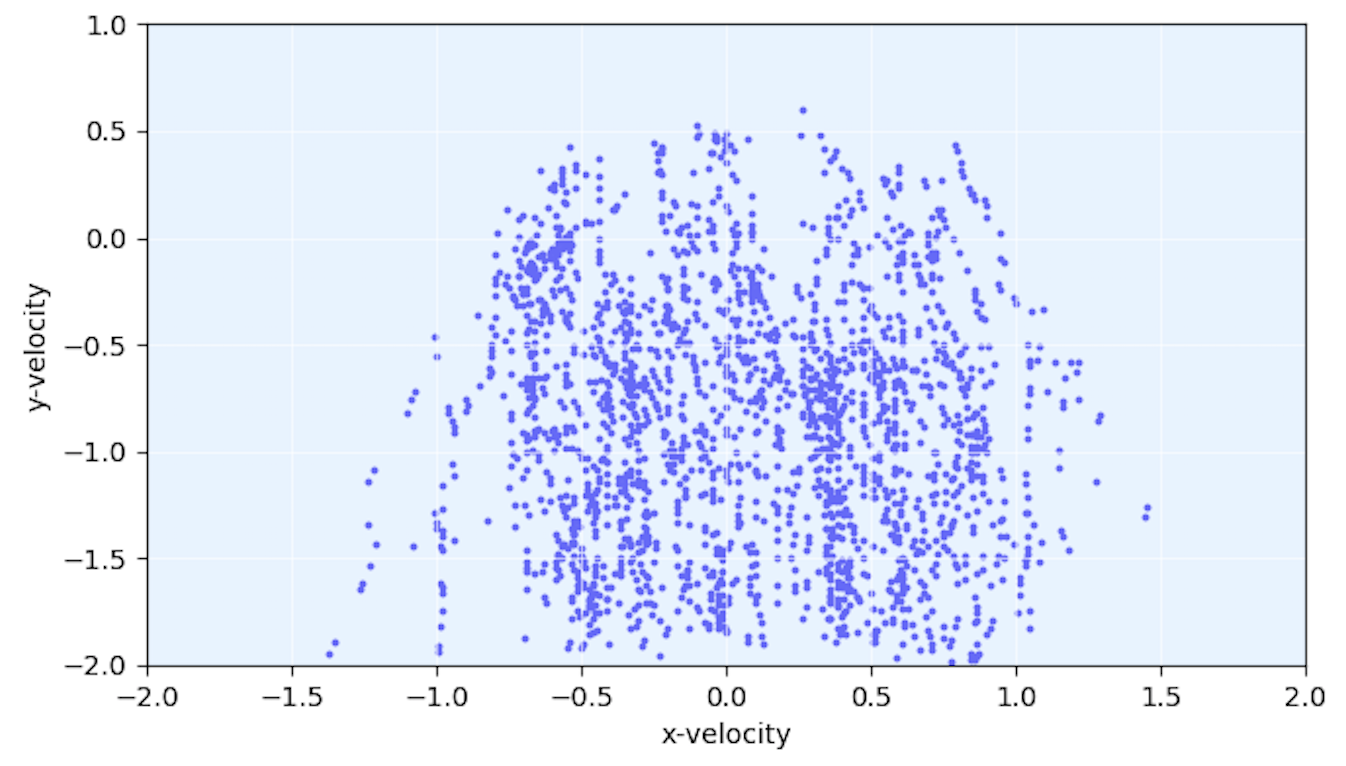}}\label{fig:f2}}
\caption[Comparing explored states in Lunar Lander for different exploration methods]
{Scatter plots of explored states in Lunar Lander after 100 episodes of only exploration. The state has 8 dimensions in this environment. Only x-velocity and y-velocity are shown. Our method didn't show any noticeable difference in the range of explored states. The best result out of 3 independent runs is plotted for each method. \textbf{(a)} Explore with random action. \textbf{(b)} Pick an action that leads to a state which has least similarity (measured by a Gaussian kernel) with recently visited states. \textbf{(c)} Our proposed algorithm for exploration.}
\label{lunar_state}
\end{figure}

This is why our algorithm did not perform better than baselines in the Lunar Lander environment. Our dynamics predictor network fails to predict the next state with high accuracy, and it's clear from Figure \ref{lunar_state} that the explored states do not follow a Gaussian distribution.

In addition, our method is prone to high dimensionality in state space. Firstly, there is a high computation cost to fit a multivariate Gaussian on high dimensional data. Secondly, numerical issues may become more likely when dimension is higher. For example, if certain dimensions of the state vector always has the same value, it will result in a singular covariance matrix.

\textbf{Future work and extensions:}\\
Instead of fitting a multivariate Gaussian to recently visited states, one can adopt a distribution that fits the observed states better. This can improve accuracy of assigning probability to a given state, increasing the chance of finding a rarer state. Our exploratory action is chosen by a one-step planning. However, if the dynamics network can predict several steps ahead with high accuracy, one can instead perform an N-step planning to pick an action that maximizes the chance of finding a rare state N steps into the future. This can be effective for environments where reaching certain states requires temporally extended planning. 

\section{Conclusion}
In this paper, we proposed DQN with model-based exploration, an improved DQN algorithm that utilizes the environment dynamics to guide exploration. We demonstrated that it outperformed the original DQN on the classic environment with sparse rewards, Mountain Car. Our algorithm was able to explore a wider range of states, and increased the learning speed. However, given the strong assumptions required, our method's effectiveness is limited to certain types of environments. For example, our experiments showed that it did not perform better than the baseline algorithms in the Lunar Lander environment, where the recently visited states are not normally distributed. We presented several ways to extend and or improve our method to solve more diversified set of environments.

\textbf{Acknowledgments.}\\
We used the following third party code: \\
\textbf{(1)} \textit{Deep Q-Learning with Keras and Gym} \cite{dqn_tutorial} as our starter code for original DQN implementation. \\
\textbf{(2)} \textit{Reinforcement learning methods and tutorials} \cite{deepRLMorvan} for our Monte Carlo Policy Gradient baseline.\\
\textbf{(3)} \textit{OpenAI gym} \cite{openai-gym} for simulated environments: Mountain Car and Lunar Lander.

\small

\bibliographystyle{unsrt}
\bibliography{references}

\appendix
\section*{Appendix}

\begin{table}[h]
  \caption{Hyper-parameters for DQN with Model-Based Exploration (Mountain Car)}
  \label{table-dqn}
  \centering
  \begin{tabular}{ll}
    \toprule                   \\
    Hyper-parameters          & Value \\
    \midrule
    $\epsilon$\textbf{ minimum }         & 0.01     \\
    $\epsilon$\textbf{ decay }          & 0.9995     \\
    \textbf{Reward discount}           & 0.99     \\
    \textbf{Learning rate (Q-network)}             & 0.05     \\
    \textbf{Learning rate (Dynamics network)}             & 0.02     \\
    \textbf{Target Q-network update interval}             & 8     \\
    \textbf{Initial exploration only steps}             & 10,000     \\
    \textbf{Minibatch size (Q-network)}             & 16    \\
    \textbf{Minibatch size (dynamics predictor network)}             & 64    \\
    \textbf{Number of recent states to fit probability model}       & 50    \\
    \midrule
    
    \multicolumn{2}{c}{}                  \\
    \multicolumn{2}{c}{Q-Network (Fully Connected)}                   \\
    \cmidrule{1-2}
    
    \textbf{Loss}                   & mean squared error \\
    \cmidrule{1-2}
    
    \multicolumn{2}{l}{Hidden Layer 1}                   \\
    \cmidrule{1-1}
    \textbf{Units}             & 48    \\
    \textbf{Activation}              & ReLU  \\
    \textbf{Initial Weights}                  & glorot uniform\\
    
    \midrule
    
    \multicolumn{2}{c}{}                  \\
    \multicolumn{2}{c}{Dynamics Predictor Network (Fully Connected)}                   \\
    \cmidrule{1-2}
    
    \textbf{Loss}                   & mean squared error \\
    \cmidrule{1-2}
    
    \multicolumn{2}{l}{Hidden Layer 1}                   \\
    \cmidrule{1-1}
    \textbf{Units}             & 24    \\
    \textbf{Activation}              & ReLU  \\
    \textbf{Initial weights}                  & glorot uniform\\
    
    \cmidrule{1-2}
    \multicolumn{2}{l}{Hidden Layer 2}                   \\
    \cmidrule{1-1}
    \textbf{Units}             & 24    \\
    \textbf{Activation}              & ReLU  \\
    \textbf{Initial weights}                  & glorot uniform\\
    
    \bottomrule
  \end{tabular}

\end{table}

\begin{table}[h]
  \caption{Hyper-parameters for Policy Gradients}
  \label{table-hyperparams}
  \centering
  \begin{tabular}{ll}
    \toprule                   \\
    Hyper-parameters          & Value \\
    \midrule
    \textbf{Learning rate}             & 0.02     \\
    \textbf{Reward discount}           & 0.995     \\
    
    \midrule
    
    \multicolumn{2}{c}{}                  \\
    \multicolumn{2}{c}{Neural Network (Policy)}                   \\
    \cmidrule{1-2}
    
    \textbf{Loss}                   & Softmax with cross entropy \\
    \cmidrule{1-2}
    
    \multicolumn{2}{l}{Layer 1}                   \\
    \cmidrule{1-1}
    \textbf{Units}             & 10    \\
    \textbf{Activation}              & tanh  \\
    \textbf{Initial weights}                  &$\mu=0$, $std=0.3$ \\
    \textbf{Initial bias}                    & 0.1 \\
    
    \cmidrule{1-2}
    \multicolumn{2}{l}{Layer 2}                   \\
    \cmidrule{1-1}
    \textbf{Units}             & dimension of action space    \\
    \textbf{Activation}              & None  \\
    \textbf{Initial weights}                 &$\mu=0$, $std=0.3$ \\
    \textbf{Initial bias}                    & 0.1 \\
    
    \bottomrule
  \end{tabular}
\end{table}

\end{document}